\documentclass[preprint,12pt]{elsarticle}

\usepackage{amssymb,graphicx,float,makeidx,amsmath}
\usepackage{amsmath,amsfonts,latexsym,cancel,float}
\usepackage{url}
\usepackage{paralist}
\usepackage{algorithm}
\usepackage{algorithmic}
\usepackage{subcaption}
\usepackage{amsthm}
\usepackage{colordvi}

\usepackage{epstopdf}

\newtheoremstyle{example*}
  {0.2cm}{0cm}
  {\normalfont}
  {0cm}
  {\itshape}{ }
  {0cm}
  {\thmname{#1}. }

\theoremstyle{example*}
\newtheorem{example*}{Example}

\newtheoremstyle{definition}
  {0.2cm}{0.2cm}
  {\itshape}
  {0cm}
  {\bfseries}{}
  {0.4cm}
  {\thmname{#1}\thmnumber{ #2} \thmnote{ (#3). }}

\theoremstyle{definition}

\newtheoremstyle{example}
  {0.2cm}{0cm}
  {\normalfont}
  {0cm}
  {\itshape}{ }
  {0cm}
  {\thmname{#1}\thmnumber{ #2}. \thmnote{ #3}}

\theoremstyle{example}

\newcommand{\dfn}{\stackrel{\mbox{\tiny def}}{=}}

\newcommand{\gram}{\mathrm{gram}}

\newcommand{\QQT}{\mathbf{Q}\times \mathbf{Q}^\intercal}
\newcommand{\QTQ}{\mathbf{Q}^{\intercal}\times \mathbf{Q}}
\newcommand{\zmax}{z_{\max}}

\newcommand{\etal}{~\emph{et al.}}
\newcommand{\eg}{\emph{e.g.}}

\begin{document}

\begin{frontmatter}



\title{On the Phase Transition of Finding a Biclique in a larger Bipartite Graph}


\author{Roberto Alonso,\corref{cor1} Ra\'ul Monroy and Eduardo Aguirre}
\ead{roberto.alonso@itesm.mx, raulm@itesm.mx, eduardo.aguirre@itesm.mx}


\address{School of Science and Engineering\\Tecnol\'ogico de Monterrey, Campus Estado de M\'exico\\Carr.~lago de Guadalupe Km 3.5, Atizap\'an, Estado de M\'exico, M\'exico}

\begin{abstract}
  We report on the phase transition of finding a complete subgraph, of
  specified dimensions, in a bipartite graph. Finding a complete subgraph in a
  bipartite graph is a problem that has growing attention in several
  domains, including bioinformatics, social network analysis and domain
  clustering. A key step for a successful phase transition study is
  identifying a suitable order parameter, when none is known.  To this
  purpose, we have applied a decision tree classifier to real-world instances
  of this problem, in order to understand what problem features
  separate an instance that is hard to solve from those that is not.
  We have successfully identified one such order parameter and with it
  the phase transition of finding a complete bipartite subgraph of
  specified dimensions. Our phase transition study shows an
  easy-to-hard-to-easy-to-hard-to-easy pattern. Further, our results indicate
  that the hardest instances are in a region where it is
  more likely that the corresponding bipartite graph will have a
  complete subgraph of specified dimensions, a positive answer. By
  contrast, instances with a negative answer are more likely to appear in
  a region where the computational cost is negligible. This behaviour is remarkably 
  similar for problems of a number of different sizes.

\end{abstract}

\end{frontmatter}


\section{Introduction}
\label{sec:intro}

In 1991, Cheeseman\etal~\cite{Cheeseman} showed that, for any
NP-complete problem, there exists a \emph{phase transition} that
separates easy instances from hard ones, and that this phase
transition can be found as one varies an \emph{order
  parameter}\footnote{Following standard convention, we use ``order
  parameter'' to refer to a parameter that controls the complexity of
  finding a solution to a given problem instance~\cite{Cheeseman},
  instead of using the more appropriate term \emph{control
    parameter}~\cite{Gent95phasetransitions}.} around one or more
critical values. Since then phase transition studies have been
conducted for a number of NP-complete problems,
see~\eg~\cite{Hogg19961,Mammen97anew,Frank,Gent94thehardest}. This is
because phase transition helps identifying key instances of a problem
so as to build a benchmark set, with which one can perform a
statistically significant comparison amongst different methods that
attempt to solve that problem. Furthermore, the phase transition value
can also be used to determine when a complete method is likely to
succeed (or not) in finding a solution to a problem instance in
reasonable time.

This paper studies the phase transition of finding a complete
bipartite subgraph, with specified dimensions, in a bipartite graph.
A \emph{bipartite graph} is a graph with two distinguished, disjoint
sets of vertices, $U$ and $V$, such that edges connect elements in $U$
to elements in $V$. A \emph{complete bipartite subgraph} is a
bipartite graph where every element of $U'\subseteq U$ is connected
with every element of $V'\subseteq V$. Henceforth, we will use
\emph{biclique} to refer to a complete bipartite subgraph. Interest in
finding bicliques inside a larger bipartite graph has started to gain
growing attention in bioinformatics (see~\eg~\cite{zhang, Mushlin,
  Sanderson}) where researchers have proposed several algorithms to
compute bicliques.  Particularly, the work of
Zhang\etal~\cite{zhang} has recently reported an improvement, with
respect to other methods, in enumerating all the bicliques in a graph
in a real-world dataset. However, finding bicliques arise naturally in
other contexts,~\eg~IoT, social network analysis, document clustering,
and privacy, amongst others.

Our motivation to study this problem originates in detecting anomalies
on Domain Name System (DNS) traffic. DNS activity when observed over a
\emph{time window}, can be represented as a bipartite graph, where $U$
is a set of IP addresses, and $V$ is the set of URLs that these IPs
have queried for. DNS traffic forms bicliques, since people tend to
visit common websites. Then, as shown by Alonso~\cite{alonso-phd}, the
number and structure of such bicliques are severely broken apart upon
an abnormal event, \eg~a denial of service attack.

For the problem of finding a biclique of specified dimensions, our
phase transition study shows an easy-to-hard-to-easy-to-hard-to-easy
pattern. A critical value occurs when the ratio of the maximal
biclique (see Section~\ref{sec:max-biclique} below) to the
cardinality of the set $V$ is roughly 3/8. There, any problem
instance would be computationally expensive but with a 98\%
probability of being solvable (see
Section~\ref{sec:phaseTransitionConcepts}). By contrast, when this
ratio tends to either zero or one, dealing with a problem instance is
negligible, being insoluble in the former case, and solvable in the
latter one.

\paragraph {Paper overview} The rest of the paper is structured as
follows. We first introduce general knowledge and formulate the
problem of finding a biclique,
Section~\ref{sec:model}.  Next, we outline how to conduct a phase
transition study, Section~\ref{sec:phaseTransitionConcepts}, and then
introduce our experimental methodology for determining a suitable
order parameter, Section~\ref{sec:orderParameter}.  Then, we present
an algorithm to compute a biclique, which applies
backtracking and a black list in order to eagerly discard vertices
that cannot form part of large biclique,
Section~\ref{sec:step3PT}. Finally, in
Section~\ref{sec:Step4phaseSGC}, we report on the phase transition of
this problem, and, in Section~\ref{sec:conclusions}, on the
conclusions drawn from our investigations.

\section{Bicliques}
\label{sec:model}

First, we shall introduce the symbols used through this document by
defining the problem of finding a biclique in a larger bipartite graph. Second,
we shall present the decision version of the problem under study,
namely finding a biclique with specified dimensions
in a larger bipartite graph. Third, we will present a way to compare the
bicliques. Lastly, we introduce the gram matrix which allows to
efficiently determine whether a maximal biclique
exists. 

\subsection{Bipartite Graphs and Complexity of Finding a Biclique}

More formally, let $G = (U,V,E)$ be a bipartite graph, with disjoint
sets of vertices, $U$ and $V$, and such that for every edge $(u,v)\in
E$, we have that $u\in U$ and $v\in V$. A biclique in $G$ is a subset
of the vertex set, we denote $g_G(U', V')$, such that $U'\subseteq U$,
$V'\subseteq V$, and such that for every $u\in U'$, $v\in V'$ the edge
$(u,v)\in E$.

The complexity of finding a biclique was initially discussed in the
work of Yannakakis\etal~in~\cite{Yannakakis:1978:NEN:800133.804355}.
There, authors have proven that finding a biclique with the
restriction that $|U'|=|V'|$ is NP-complete. Later, works
like~\cite{Dawande:2001:BMC:509217.509229,Peeters2003651} have proven
that the NP-completeness of finding a biclique also holds for other
restrictions, such as specifying a maximum number of edges, a maximum
number of vertices, or specifying a maximum edge weight. Particularly,
the version of the problem for which we demand $|U'|=|V'|$ is called
\emph{balanced biclique}~\cite{Yannakakis:1978:NEN:800133.804355}, and
that where $|U'|=t$ and $|V'|=z$, for given $t$ and $z$, is called
\emph{exact node cardinality
  decision}~\cite{Dawande:2001:BMC:509217.509229}. Recently, Alonso
and Monroy~\cite{alonso2} have proven that finding a biclique such that
$|U'|\geq t$ and $ |V'|\geq z$ vertices, for given $t$ and $z$, is
also NP-complete.  They have also shown that even if we try to prove
that every element in $U$ is in a biclique with at least two elements
of $V$ remains in the class NP. Authors called these problems Social
Group Commonality (SGC) and 2-SGC, respectively.

\subsection{Finding a Biclique with Specified Dimensions $t$ and $z$: a Decision Problem}
\label{sec:sgc-dp}
Now, the decision problem of finding a biclique with specified dimensions is defined as follows:
\begin{compactitem}
\item[]INSTANCE: A bipartite graph $G=(U,V,E)$, two positive integers,
  $t$ and $z$.\vspace*{0.05in}
\item[]QUESTION: Is there a biclique in $G$, $g_G(U',V')$, with
  $|U'|\geq t$ and $|V'|\geq z$?
\end{compactitem}
\vspace*{0.1in} 
Given that a graph may contain several bicliques, in what follows we
present a way to compare them. This formulation will be used later in
the phase transition study.

\subsection{Size-/Weight-Maximal Biclique}
\label{sec:max-biclique}

Let $G=(U,V,E)$ be a bipartite graph and let $g_G(U',V')$, with
$U'\subseteq U$ and $V'\subseteq V$, be a biclique in $G$. Then, we call
$|U'|$ and $|V'|$ the \emph{weight} and the \emph{size} of $g_G(U',V')$,
respectively. Further, let $\mathcal{G}_G$ denote all the possible
bicliques in $G$. Then, a biclique $g_G(U',V')\in \mathcal{G}_G$ is called
\emph{weight-maximal} (respectively, \emph{size-maximal}) if there is
not a $g_G(U'',V'')\in\mathcal{G_G}$ such that $|U'|<|U''|$
(respectively, $|V'|<|V''|$). 


A bipartite graph $G$ can be succinctly represented by means of an adjacency matrix $\mathbf{Q}$. From the gram matrix of $\mathbf{Q}$ it is possible to get insights about the graph, as we will show below.
\subsection{The Adjacency Matrix and the Gram Matrix}

Let $G = (U,V,E)$ be a bipartite graph. We use $I,J,\dots$ stand for indexing sets, and write $U_I = \{u_i:u\in U,i\in I\}$ to denote the nodes in $U$, indexed by $I$; likewise, $V_J = \{v_j:v\in V,j\in J\}$ denotes the nodes in $V$, indexed by $J$. Then,  
an \emph{adjacency matrix}, $\mathbf{Q}$ is such that $\mathbf{Q}_{(i,j)} = 1$ implies that there is an edge $(u_i,v_j)\in E$. 

Now, we can compute the gram matrix denoted $\gram(\mathbf{Q})$ and given by $\QQT{}$. In particular, $\gram(\mathbf{Q})$ is symmetric, and such that
the lower (respectively, upper) triangular matrix contains information
about \emph{all} the distinct bicliques with weight two,
including the one that is size-maximal. Notice that
$\gram(\mathbf{Q})_{(k,l)} = n$, $k\neq l$, implies that the graph contains
a biclique with weight two and size $n$, involving the participation of
vertices $u_k$ and $u_l$.  The main diagonal of this matrix enables us
to determine the number of adjacent vertices of $u_k$, since $\gram(\mathbf{Q})_{(k,k)}=n$ implies that
vertex $u_k$ has $n$ adjacent vertices in the graph.

Complementarily, $\gram(\mathbf{Q}^\intercal)$, given by $\QTQ$,
provides valuable information about the bicliques in the graph. In particular, the lower (respectively, the upper) triangular
matrix of this matrix contains \emph{all} the distinct bicliques with size two, including the one that is weight-maximal. Here, $\gram(\mathbf{Q}^\intercal)_{(k,l)}=n$ implies that there is a biclique with size two and weight $n$, involving the use of $v_k$ and $v_l$ vertices. $\gram(\mathbf{Q}^\intercal)_{(l,l)}=n$ implies that vertex $v_l$ has $n$ adjacent vertices. 

\textbf{Remark:}~We should point out that the gram matrix of $\mathbf{Q}$ and $\mathbf{Q^\intercal}$ provide a proof of the existence of a size-maximal (respectively, weight-maximal) biclique with weight two (respectively, size two), while determining the existence of a maximal biclique with weight (respectively, size) greater than two remains NP-complete.


\section{Standard Methodology for a Phase Transition Study}
\label{sec:phaseTransitionConcepts}

Phase transition is a means of selecting problem instances that are
typically hard, and hence provide a fair basis for comparison of
different algorithms. A phase transition, separating easy instances
from hard ones, appears as one plots the expense of finding a solution
to a problem instance against an order parameter. Interestingly, it
often coincides to that area where the problem, stated as a decision
problem, changes from having a YES-solution (\emph{solvable}) to one
having not (\emph{insoluble}). The term is used in an analogy to the
Physics phenomena: after a phase transition, a material dramatically
changes its properties, \eg~from liquid to solid.

Some problems have been found to show an easy-to-hard-to-easy
complexity pattern (\eg, travelling
salesman~\cite{Cheeseman,Gent96thetsp}): the cost of finding a
solution increases at first, but then decreases later on to small
values back again. Others (\eg~constraint
satisfaction~\cite{Gent95scalingeffects}) have been found to have
similar cost, regardless of the size of the instance, as long as the
instance is scaled. 
\emph{Scaling} has several implications; it can be used to construct
an instance with a given probability, or a set of instances with
similar cost, and this can be done for any problem size.

Conducting a phase transition study is a four-step approach:
\begin{compactenum}
\item Select an order parameter that succinctly captures the problem
  structure. This task may not be trivial. As pointed out
  by~\cite{Cheeseman}, using a different order parameter yields a
  different phase transition.  While several NP-complete problems
  exhibit a natural order parameter, others require experimental
  evidence; \eg~Gent and Walsh~\cite{Gent96phasetransitions} used an
  annealed theory to determine an order parameter for the phase
  transition of number partitioning. In this work, we have applied a decision tree 
  classifier to real-world instances of the problem dealt with on this paper, in order to 
  understand what problem features separate an instance that is hard 
  to solve from those that is not. We have successfully identified one 
  such order parameter (see
  Section~\ref{sec:orderParameter}) and with it the phase transition of finding a 
  biclique of specified dimensions.
  

\item Collect a number of problem instances. This can be done either
  by randomly generating problem instances using the selected order
  parameter, or by collecting them from a real-world process, if any.
  For our study, we have collected over 100 thousand bipartite graphs from a real-world process. 
  \label{it:collection}

\item Select an algorithm that solves the problem, and then apply it
  on each instance of the set built from the second step; for each
  try, gather both computational expense and whether it is solvable or
  not. In our study, we have designed and applied an algorithm for this task
  (see Section~\ref{sec:step3PT}). Notice that, alternatively, in this
  step we could have used an efficient algorithm, \eg~a SAT solver;
  but then we would have to come out with a mapping from the problem of finding a biclique to SAT,
  which is beyond the scope of this paper.

\item Plot both the computational cost, and the probability of an
  instance being solvable against the order parameter. In our case,
  this probability is given by the number of instances that were
  solvable divided by the total number of instances considered, for
  each value of the order parameter.  Notice that accomplishing this
  step depends on how step~\ref{it:collection} is carried out. Had
  data been synthesized, the generation function would have to tag
  each instance with the probability of it being solvable.
\end{compactenum}

\section{On Identifying an Order Parameter for the Problem of Finding a Biclique}
\label{sec:orderParameter}

In order to identify an order parameter, we have characterized real-world
instances for which it is possible to find a biclique with
little effort, and those that cannot. To that purpose, we have applied
C4.5, which builds a decision tree from a training set containing
already classified graph samples. This tree can be separated into decision
rules, which explain what makes an instance to be one class or the
other. In what follows, we first describe our working dataset, and
then how C4.5 was applied to it to discover an order parameter.

\subsection{Dataset Construction}

Our dataset, including both training, test and validation, has been built out
of real-world activity, namely a Domain Name System (DNS) resolution process. Roughly, a DNS resolution process is about an agent querying for a domain so as to translate them into an IP address. Then, by observing $w$ DNS processes, we have constructed a graph $G = (U,V,E)$, where the set $U$ denotes IP addresses, the set $V$ denotes URLs and $(u,v)\in E$ represents the action of agent $u$ over domain $v$, namely the process of translation. From this bipartite graph it is possible to observe bicliques (since we tend to visit the same websites), as shown in~\cite{alonso-phd}.

This kind of real-world process usually involves a large and dynamic graph that follows a behaviour similar to a free scale network, so in our case we can end up with a 
graph comprising in average 600K vertices per day. Because decision trees on large data sets can be time consuming and hard to store in memory, we have decided to construct more manageable graphs in the following way:

First we arbitrarily picked five days of DNS 
traffic. Then, we randomly sampled $w$ DNS processes, and constructed the adjacency matrix $\mathbf{Q}$. We should point out that given the nature of the real-world process, repetition may occur (\eg~an IP may ask for the same domain more than once) however this does not get in the way of constructing the graph or computing the biclique. We constructed graphs so as to attain a set amounting 40\% of each picked day. We repeated this procedure for $w$ from 50, to 150 in steps of 25. After this, step we end up with a collection for each of the selected $w$.

For each graph in the collection, we proceeded as follows.
First, we preemptively applied a brute-force algorithm in order to find a size-maximal biclique. 
If it could be solved in less than 20
seconds, we labelled it \emph{EASY}; otherwise, we gave up solving it
and labelled it \emph{HARD}.  Then, we inserted in the final dataset a
tuple containing a feature vector (described below) representing the graph, and
the associated label.  We finally split this dataset, forming
the \emph{training set} (comprising 70\% of the data) and the
\emph{validation set} (comprising the remaining 30\%).

\subsection{Features Used to Characterize a Graph}

We now show the feature vector representing a graph. We
insist that in the selection of all these features, we were driven by
determining an order parameter, and that they all capture the
likelyhood of an instance being HARD. These features are:
\begin{compactitem}
\item $|U|$: the cardinality of $U$.

\item $|V|$: the cardinality of $V$.

%

\item $|E|$: the cardinality of $E$.

\item An estimation of the total number of object combinations that
  need to be attempted to search for bicliques, denoted
  $\mathrm{comb}(G)$. Take a graph $G$, compute $\gram(\mathbf{Q})$,
  and then look for the three highest values of the lower triangular
  gram matrix and multiply these values.


\item The ratio $(|U||V|)/w$, we call the
  \emph{social degree} of $w$.
  




\item The weight $U'$ of a weight-maximal biclique in $G$.  
\item The size $V'$ of a size-maximal biclique in $G$. 

  
\item The number of 2-weight bicliques, computed from
  $\gram(\mathbf{Q})$. 


\item And, likewise, the number of 2-size bicliques.

\end{compactitem}

\subsection{Construction of the Classifier}
\label{sec:orderParameterExpSetting}

We have built seven classifiers, one per each selected $w$ considered in our
dataset. The rationale behind this design decision is to observe
whether, and if so how, the number of observations is part of the order parameter.
We have built each classifier using ten-fold cross-validation.
Roughly, we first randomly picked 90\% of the training set. Then, we
obtained a classification tree from these data, using C4.5, as
implemented in Weka~\cite{datamining}. Second, we tested the tree on
the remaining 10\% of the instances.  Third, we repeated this
procedure 10 times.  Finally, we selected the best classification tree
and validated it on the test set. The corresponding results are
reported on below.

\subsection{An Evaluation of the Classification Tree Performance}

Table~\ref{tbl:falsepn} shows the false positive rate and the output
by our classifier, for various collections. The \emph{false positive 
rate} (FPR) is the rate at which the classifier mistakes a HARD
instance to be EASY, and the \emph{false negative rate} (FNR) the
other way round. There usually is a trade-off between these rates.
Notice how the FPR grows as the number of observations does. This is explained by
the larger the number of observations, the larger the proportion of
instances labelled HARD.  This implies that in the dataset, both
training and validation, classes are not balanced.  Thus, it is more
likely that a HARD instance is wrongly classified.

\begin{table}
  \centering
  \begin{tabular}{| c |c| c|}
    \hline
    Collection with $w=$ & FPR & FNR \\ 
    \hline \hline
    100    & 0.770              & 0.259               \\ 
    \hline
    150    & 5.609              & 0.199                \\ 
    \hline
    200    & 13.370            & 0.197               \\ 
    \hline
    250    & 17.5                & 0.259              \\ 
    \hline
  \end{tabular}
  \caption{Classifier outcome: FPR and FNR.}
  \label{tbl:falsepn}
\end{table}

\begin{figure}[t!]
\centering
 \includegraphics[scale =0.7] {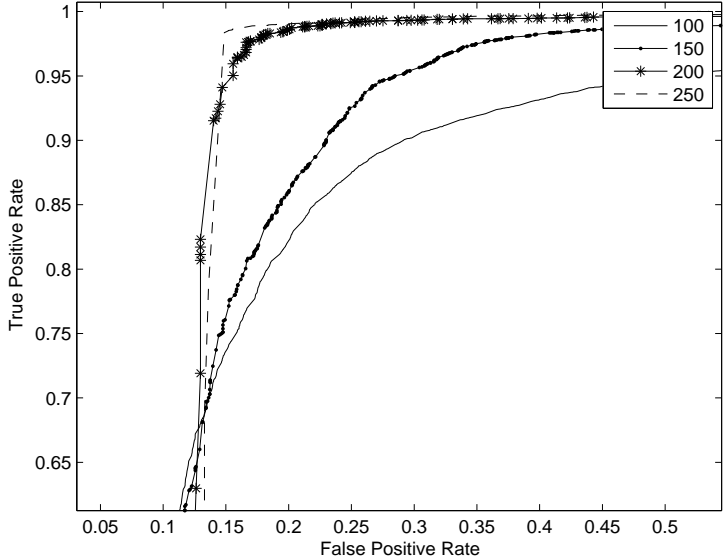} 
  \caption{ROC curves for graph collections with $w$ form 100 to 250 in steps of 50.}
\label{fig:roc}
\end{figure}

More thoroughly, we have evaluated the performance of our classifier
using Receiver Operating Characteristic (ROC) curves
(see~Figure~\ref{fig:roc}). A \emph{ROC curve} is a parametric curve,
generated by varying a threshold and computing both the FPR and the
FNR, at each operating point.  The upper and the further left a ROC
curve is, the better the classifier is.  Figure~\ref{fig:roc} shows
that our classifier is able to recognise instances. Notice that the
classifier performance improves along with the number of observations.

\begin{figure}[t!]
\centering
 \includegraphics[scale =0.7] {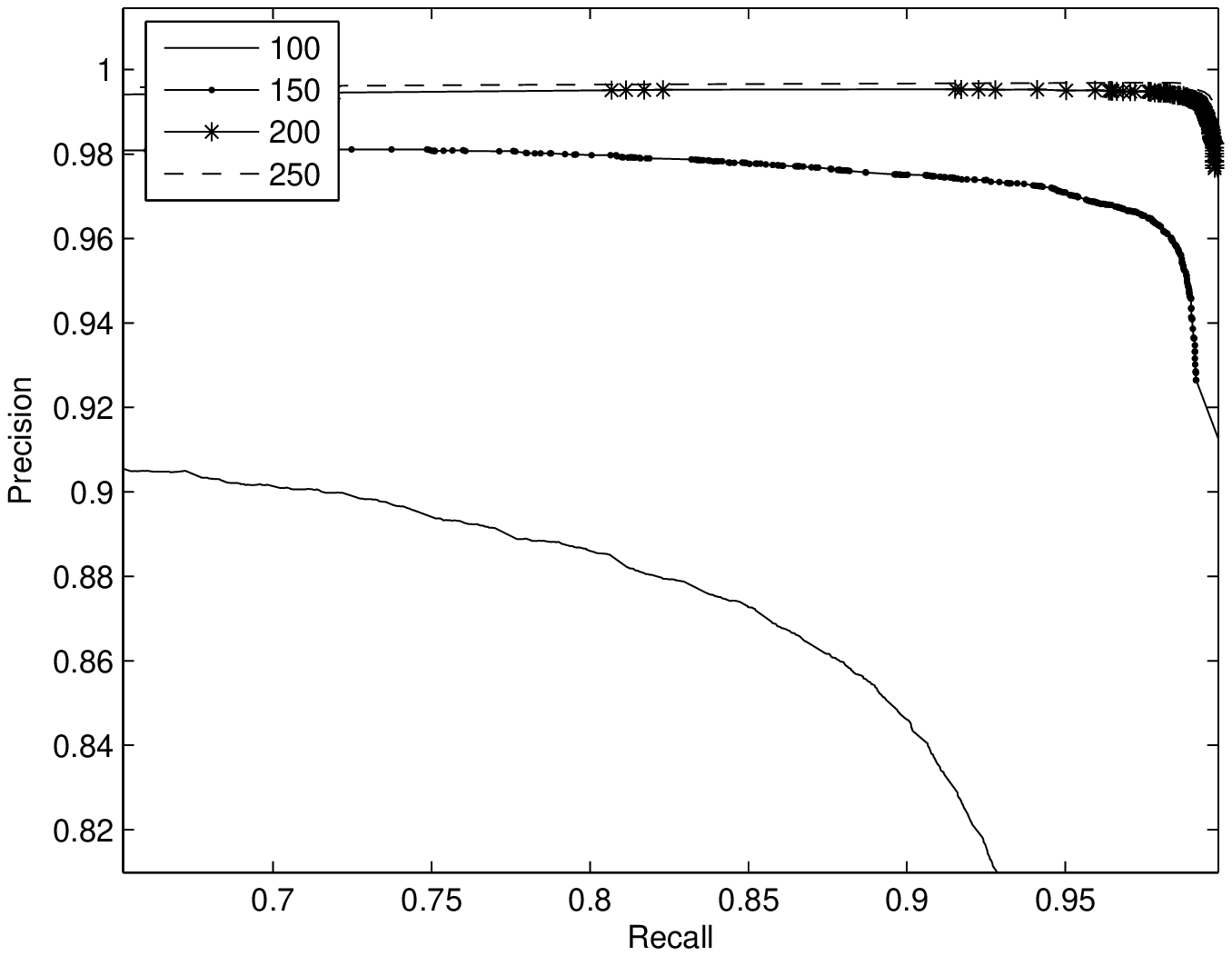} 
  \caption{Precision-Recall curves for collections with $w$ form 100 to 250 in steps of 50.}
\label{fig:pr}
\end{figure}

In order to support these results, we have plotted precision over
recall. Here, the upper and further to the right the curve is, the
better classifier is. Figure~\ref{fig:pr} shows again how the
classifier performance improves with the number of observations.  This can be
attributed to both class unbalance, and to the occurrence of a higher
proportion of HARD instances in large graphs.

\subsection{Order Parameter Discovery from the Classification Tree}
\label{sec:orderParamDisco}

\begin{figure}[ht]
  \centering
  \includegraphics[scale =0.5] {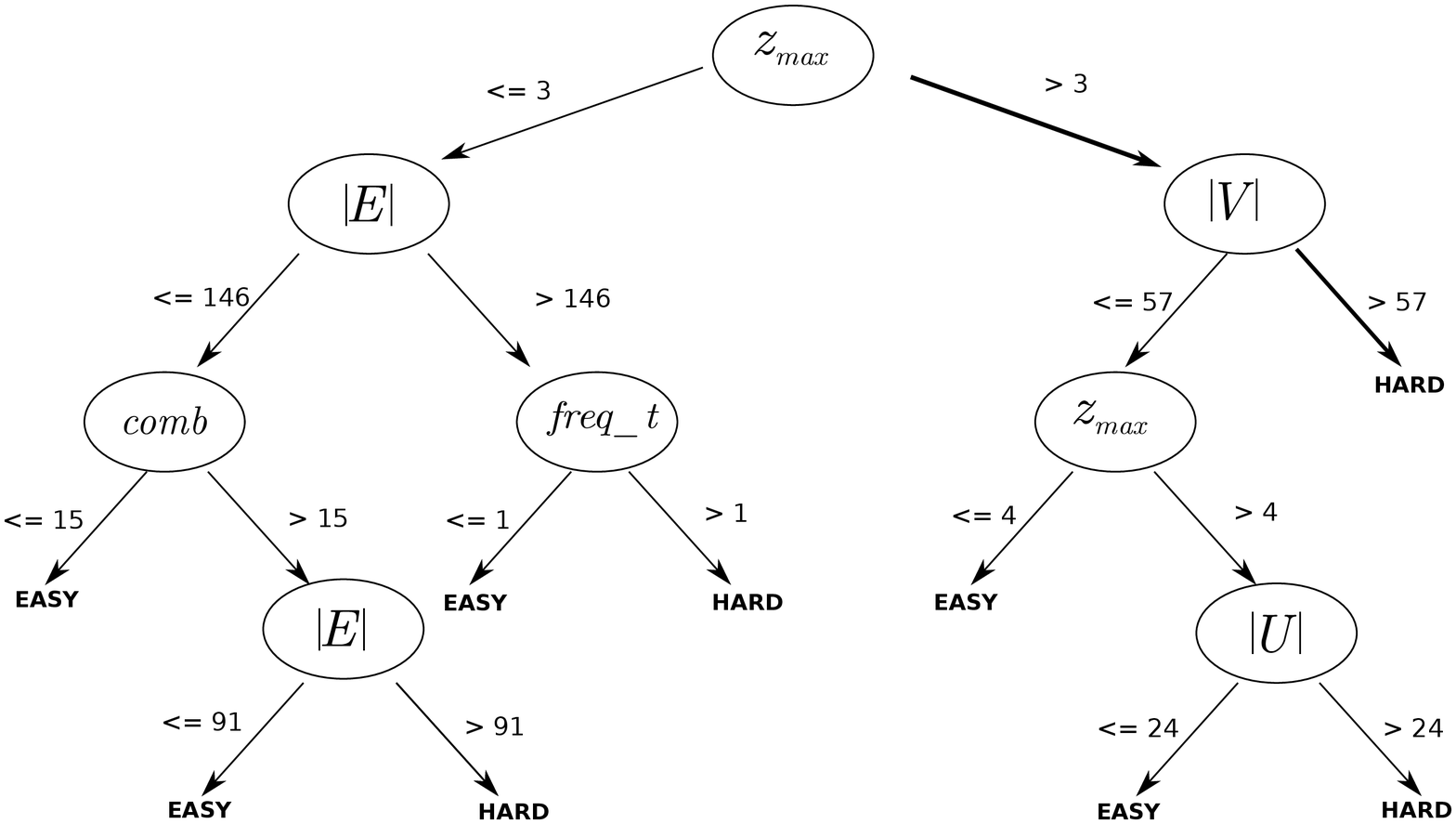}
  \caption{C4.5 classification tree for a collection with a number of observations $w = 250$.  Here, $comb$ is the number of combinations to be attempted.} 
  \label{fig:tree}
\end{figure}

The classification trees we have obtained for all datasets are
remarkably similar, regardless of the number of observations $w$.
Figure~\ref{fig:tree} displays the one for a collection of graphs with $w = 250$ observations.
In general, the rules extracted out of these classification trees show
that the features that separate a HARD instance from an EASY one are
the cardinality $|V'|$ of the size-maximal biclique (denoted $\zmax$), the number of edges in the graph $|E|$, number of 2-weight bicliques (denoted $freq_t$), the cardinality $|U|$ and the cardinality $|V|$.

\begin{figure}[ht]
  \centering
  \includegraphics[scale =0.65] {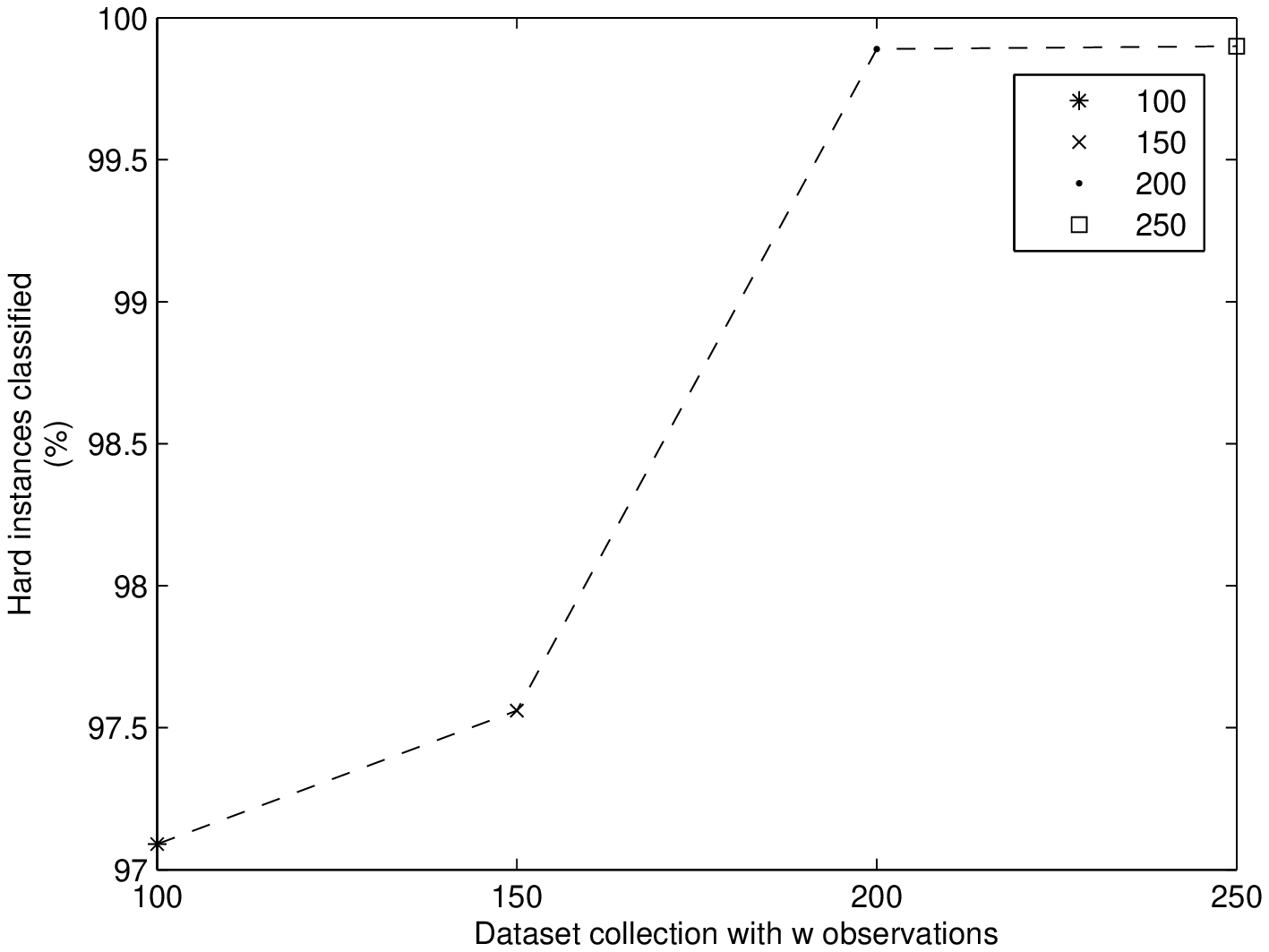}
  \caption{Proportion of HARD graphs classified correctly with the rule.} 
  \label{fig:common}
\end{figure}

We also noticed that a large number of instances of type HARD are
captured by one rule, namely: \emph{label instance HARD, if cardinality $V'$ of
  size-maximal biclique and cardinality $|V|$ are respectively greater
  than 3 and 57}.  We constructed a classifier considering this rule
only.  Figure~\ref{fig:common} shows that this classifier is able to
distinguish most of the HARD instances, regardless of the number of observations.

Considering this result, we have come up with the following order
parameter, $\pi$. Let $\zmax$ denote the cardinality of a size-maximal biclique, then $\pi\dfn\zmax/|V|$, with $\pi\in[0,1]$.
Notice that, as $\zmax/|V|\to 1$, finding a size-maximal is
computationally harder, as we might need to explore the entire search
space. By contrast, as $\zmax/|V|\to 0$ most the search
space can be pruned, since it is easy to discard the existence of
large bicliques.

\section{An Algorithm to Compute Bicliques}
\label{sec:step3PT}

To conduct our phase transition study, we have used
Algorithm~\ref{alg:algorithm},\footnote{A Perl implementation of this
  algorithm is available at \url{https://db.tt/vDG5vXh5.}} which,
given a graph and a positive integer $z$, returns a biclique with size
$z$ and maximal weight, if any, along with the computational cost
incurred. Function $adjacentTo(V)$ returns a set of
vertices $U'\in U$, where \textit{all} $u\in U'$ are adjacent to $V$. Notice that in lines 8-9 the algorithm
returns a biclique, if any;
otherwise, it returns $noSolution$. Also, notice that the biclique found by the algorithm is weight-maximal. This does not add any computational cost to the problem since our measure of computational expense, shown below, makes use of the other metric. Moreover, there is knowledge of the cardinality of a weight-maximal biclique, obtained from $\gram(\mathbf{Q}^\intercal)$, so finding a biclique larger than the weight-maximal is negligible using our algorithm.

\renewcommand{\algorithmicrequire}{\textbf{Input:}}
\renewcommand{\algorithmicensure}{\textbf{Output:}}
\renewcommand{\algorithmiccomment}[1]{// #1}
\begin{algorithm}[h!]
  \begin{algorithmic}[1]
    \REQUIRE A bipartite graph $G = (U,V,E)$ and a positive integer $z$ of the biclique being looked for.
    \ENSURE A biclique size $z$ with maximal-weight along with the
    corresponding witness, if any.
    
    \STATE $r \gets 2$
    \STATE $BlackList \gets \emptyset$ 
    \WHILE{$z \geq r$}

    	\STATE $\mathbb{V} \gets {V \choose r}$ \COMMENT{$\mathbb{V}$ is set of sets of vertices $V$}
    
    	\FORALL{$V'$ in $\mathbb{V}$} 
    		\IF {$ V' \notin BlackList$}
			\STATE $U' \gets adjacentTo(V')$    

			\IF{$|U'| \geq 2$  \AND $z = r$ }
				\RETURN $G(U',V',E)$
				\STATE \COMMENT{Return a biclique witnessed by $U'$ and $V'$}
			\ELSE
				\IF{$|U'| < 2$}
				\STATE $BlackList \gets BlackList\cup V$
				\ENDIF
			\ENDIF
			\ENDIF 
    	\ENDFOR
    	\STATE $r = r+1$
    \ENDWHILE
    \RETURN $noSolution$
  \end{algorithmic}
  \caption{Backtracking based approach to compute a biclique in a larger bipartite graph.}\label{alg:algorithm}
\end{algorithm}

Following~\cite{Gent94thesat,Gent95scalingeffects,Gent96thetsp}, we
used the number of combinations explored as a measure of computational
expense. The rationale behind this decision is it is not affected by
the hardware on which experiments are run. This is in contrast with other measures, such as time-to-solve (used~\eg~in~\cite{torres}). We
run our algorithm in over 100 thousand bipartite graphs from our dataset (described in Section~\ref{sec:orderParameter}) using several computers.
Our experimentation on this work lasted about 6 months of continuous
calculations using two computers: the first one being a Core i7 2Ghz
computer with 4GB in RAM, and the second one a two Xeon 3GHz computer
with 8GB in RAM. 

\section{The Phase Transition of Finding a Biclique in larger Bipartite Graph}
\label{sec:Step4phaseSGC}

We are now ready to present the phase transition. First,
Section~\ref{sec:phasedecision}, we present our results for the
decision problem, as introduced in Section~\ref{sec:sgc-dp}. Next, in
Section~\ref{sec:phaseoptimality}, we present an alternative phase
transition, where the problem is now turned into finding a size-maximal biclique 
with maximal weight.

\subsection{Phase Transition of the Decision Problem of Finding a Biclique}
\label{sec:phasedecision}

\begin{figure}[hbt]
\centering
\includegraphics[scale=0.6]{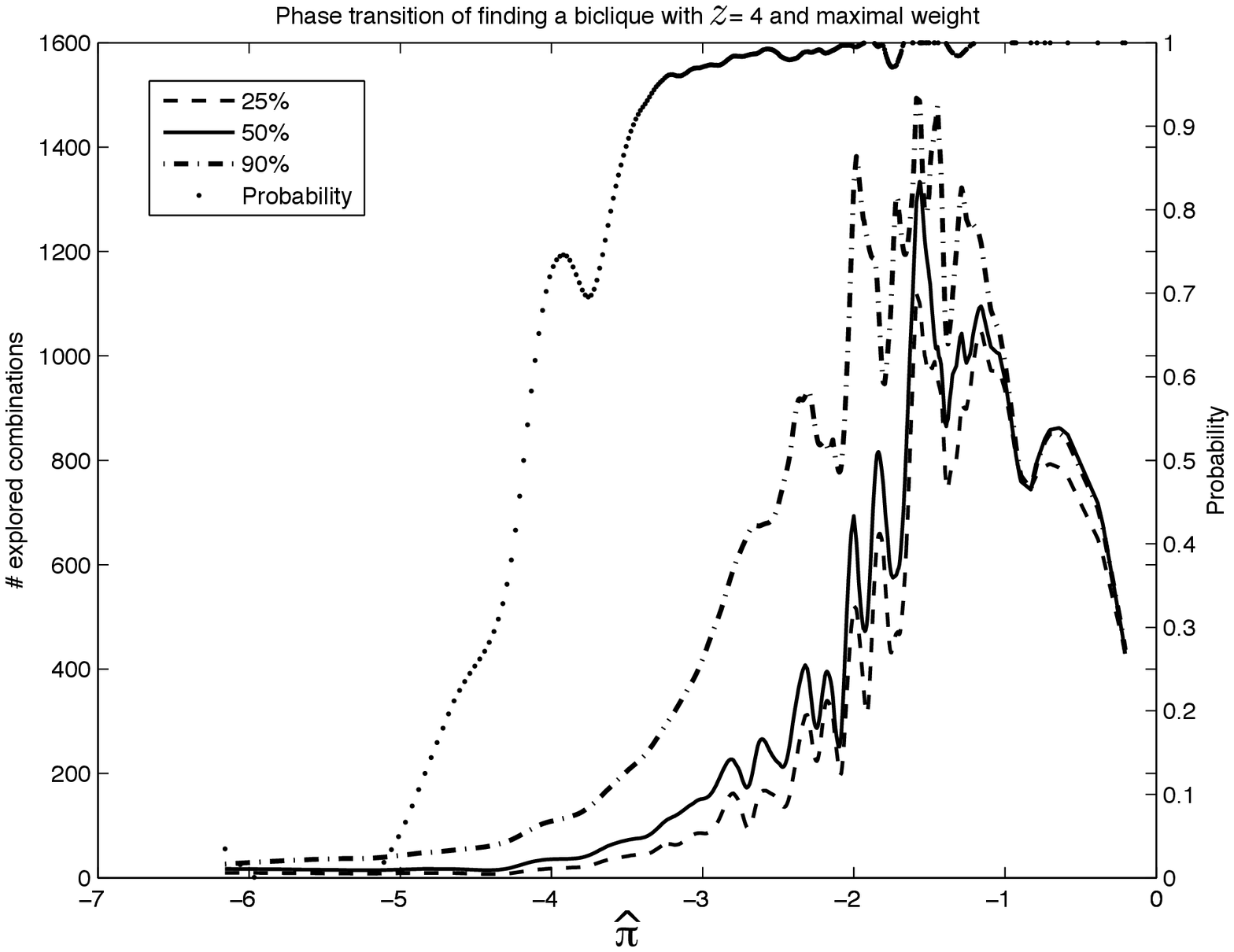}
\caption{Percentile 90\%, 25\% and median cost of finding a biclique with size
$z=4$ and maximal weight for a number of observations $w$ from 50 to 150 in steps of
  25.} 
\label{fig:percentileSize4}
\end{figure}

\begin{figure}[hbt]
\centering
\includegraphics[scale=0.6]{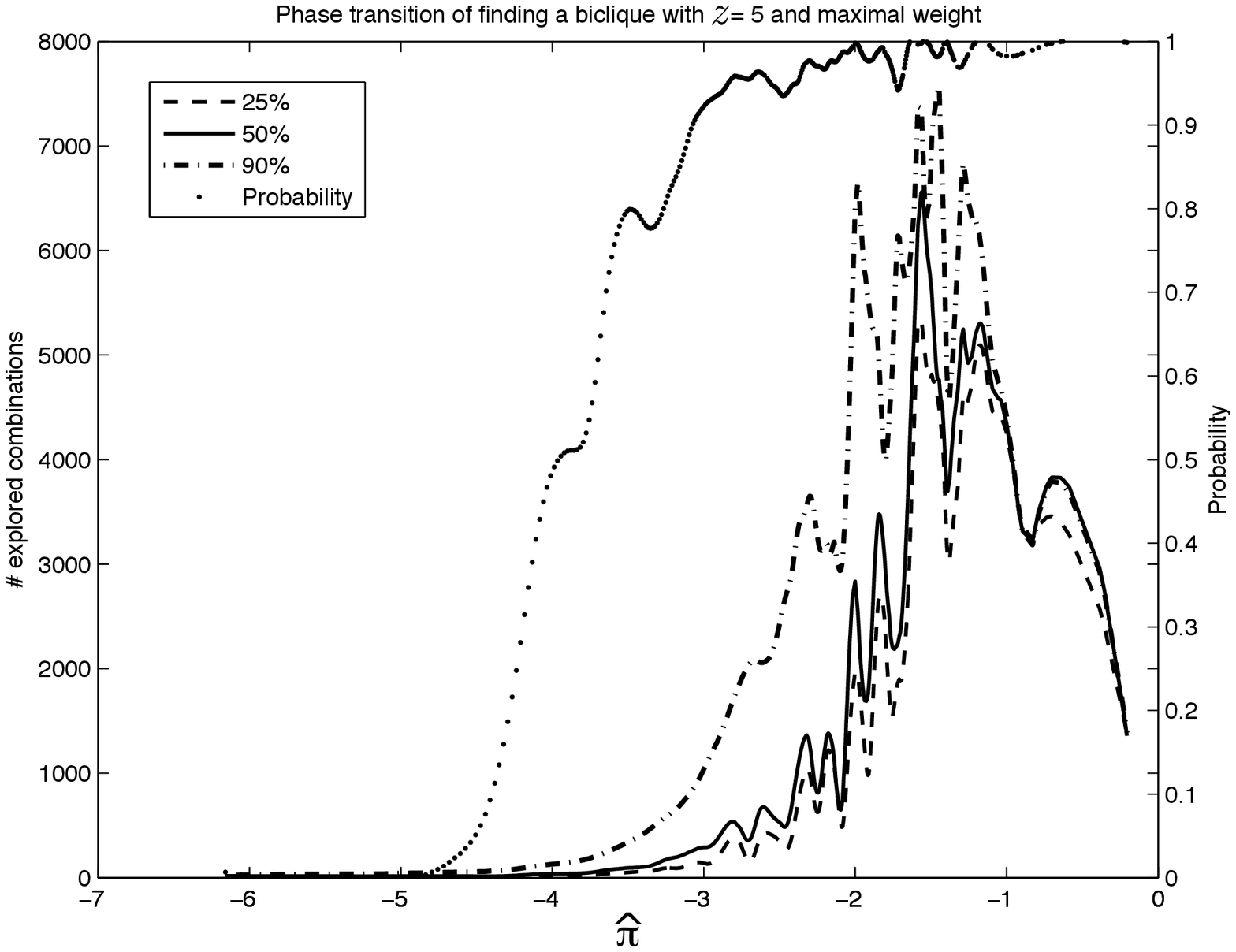}
\caption{Percentile 90\%, 25\% and median cost of finding a biclique with size
$z=5$ and maximal weight for a number of observations $w$ from 50 to 150 in steps of
  25.} 
\label{fig:percentileSize5}
\end{figure}

\begin{figure}[hbt]
\centering
\includegraphics[scale=0.6]{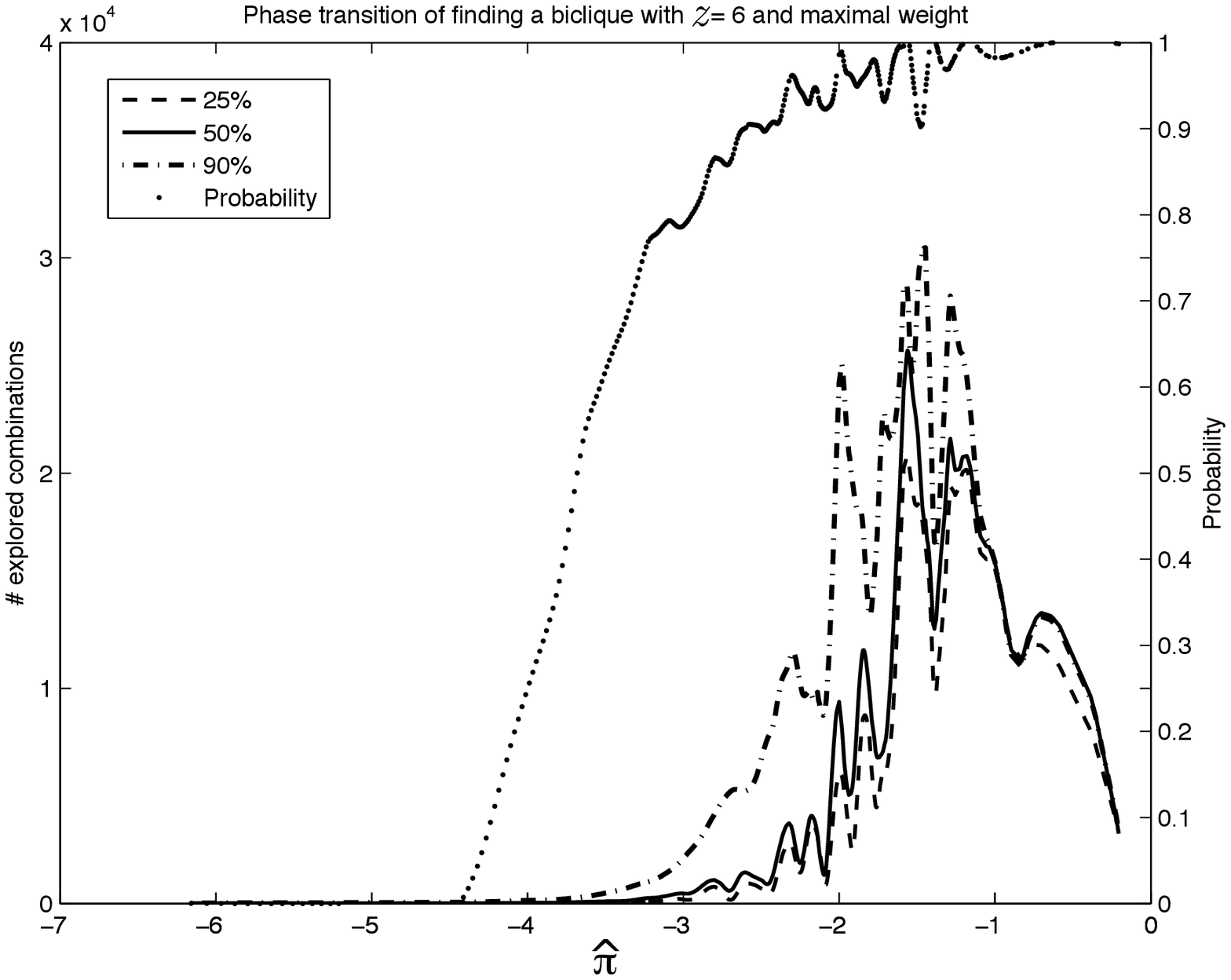}
\caption{Percentile 90\%, 25\% and median cost of finding a biclique with size $z=6$ and maximal weight for a number of observations $w$ from 50 to 150 in steps of 25.}
\label{fig:percentileSize6}
\end{figure}

\begin{figure}[hbt]
\centering
\includegraphics[scale=0.6]{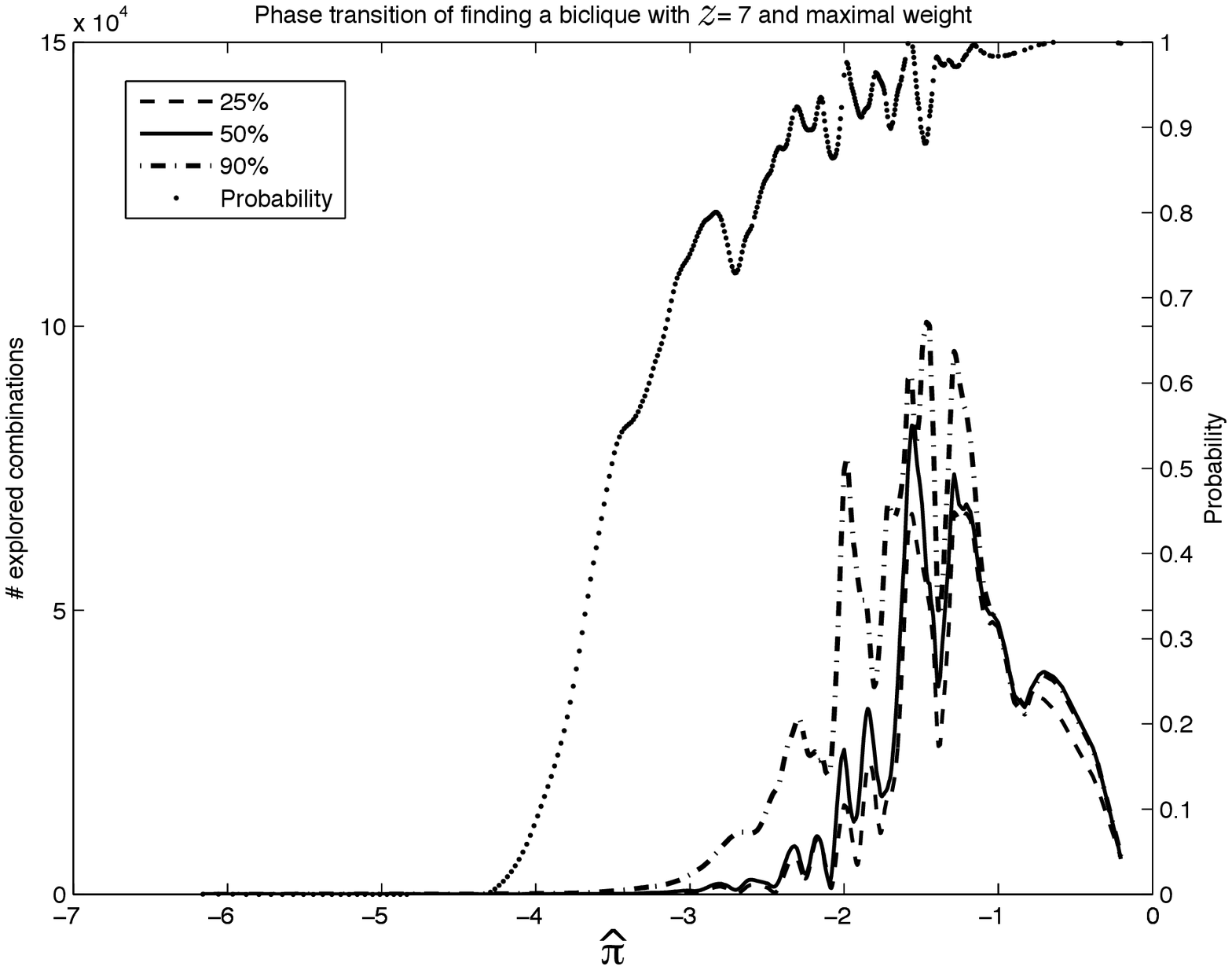}
\caption{Percentile 90\%, 25\% and median cost of finding a biclique with size $z=7$ and maximal weight for a number of observations $w$ from 50 to 150 in steps of 25.}
\label{fig:percentileSize7}
\end{figure}

\begin{figure}[hbt]
\centering
\includegraphics[scale=0.6]{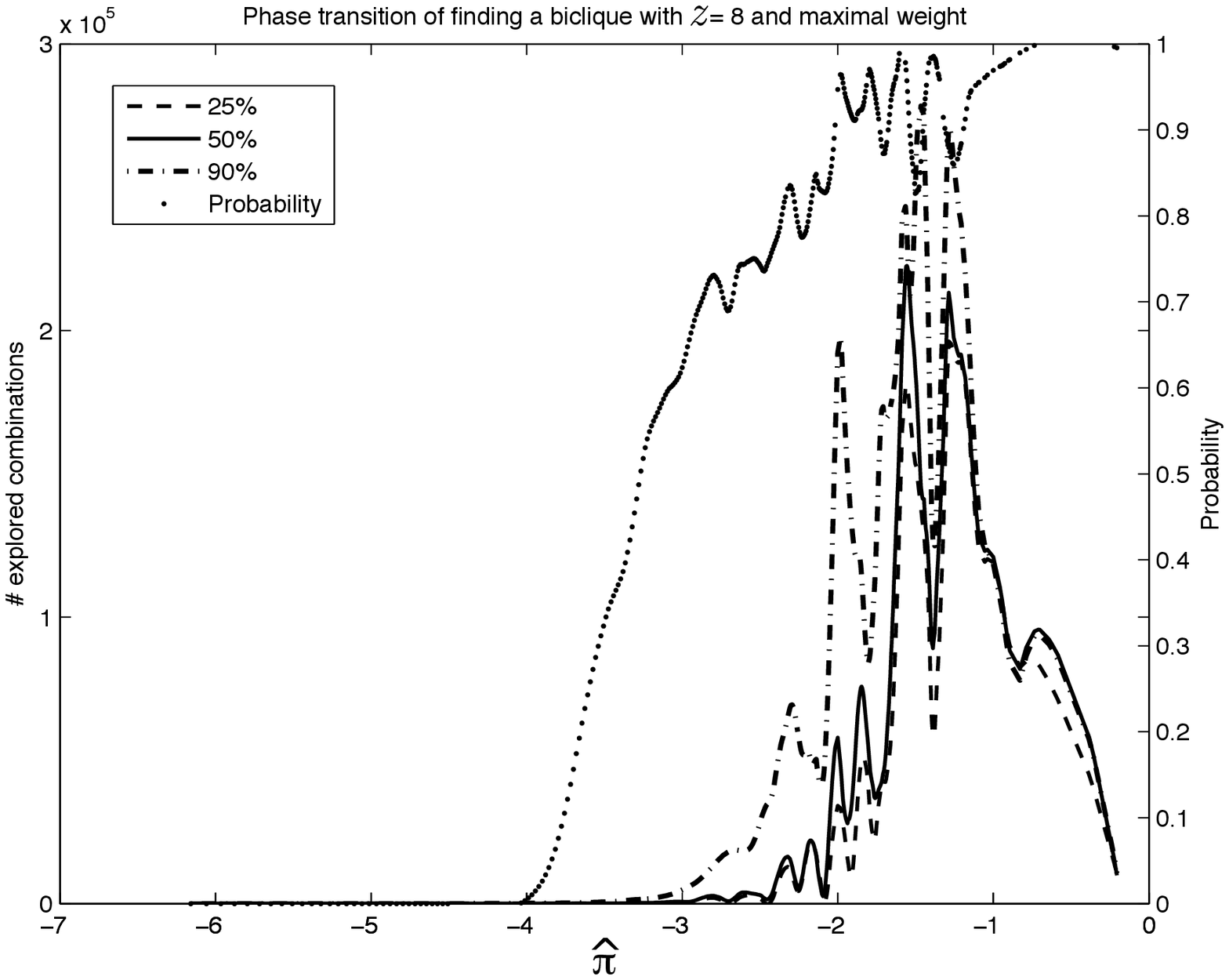}
\caption{Percentile 90\%, 25\% and median cost of finding a biclique with size $z=8$ and maximal weight for a number of observations $w$ from 50 to 150 in steps of 25.}
\label{fig:percentileSize8}
\end{figure}

Figs.~\ref{fig:percentileSize4} --- \ref{fig:percentileSize8} show the
phase transition, where we have set the size $z$, to be
equal to 4 --- 8, respectively. On each curve, we have set the weight to be maximal and combined the results for the collection with $w$ from $50, 75, \ldots, 150$ number of observations. As standard in literature, the horizontal axis represents the order parameter, in our case denoted $\pi$ and given by the size of a size-maximal biclique divided over the cardinality of $V$ in an instance, in symbols $\pi\dfn\zmax/|V|$ (see Section.~\ref{sec:orderParamDisco}). Complementarily, in the vertical axis we have plotted the computational cost, the number of explored combinations in our case, involved in finding bicliques. To display the transition better,
instead of $\pi$, we have plotted
$\widehat{\pi} = log_2(\pi)$, the order parameter in logarithmic
scale.

Figs.~\ref{fig:percentileSize4} --- \ref{fig:percentileSize8} all
adopt the following conventions. A solid line is used to denote the
median cost of finding a biclique with size $z$; a dashed line to denote
the 25th percentile (representing the less expensive instances); a dash-dotted line to denote the 90th percentile (representing
the hardest instances); and, finally, a dotted line is used to
capture the probability of an instance being solvable.

Interestingly, the curves all show the same easy-hard-easy-hard-easy pattern,
for median, 25th percentile, 90th percentile. 
Roughly, the hardest instances, showing the highest associated
computational cost, all lie at $\widehat{\pi}=-1.5$ (thus, $\pi=3/8$).
Remarkably, this cost inflexion appears even in the 25th percentile
curve. Notice how after the high inflection point,
$\widehat{\pi}=-1.5$, the computational cost decreases as
$\widehat{\pi}\leadsto 0$. This result implies that it is negligibly
cheap to find a biclique in a larger graph where a size-maximal has a size 
similar to the cardinality $V\in G$, and, that, hence, a biclique is likely
to be found. This is attributable to the adjacency matrix, which is
mostly full of one values, and it is easy to discard vertices that cannot form part of a larger biclique. Thus,
bicliques with $z\leadsto|V|$ are uncommon, but fortunately
take no time to solve.

Notice that for $\widehat{\pi}<-5$, it is negligibly cheap to realize
that an instance does not have a biclique. This result implies it takes no
effort to find a biclique, where size-maximal bicliques all have a
very small size compared to the cardinality of $V$; notice how
$\pi=1/32$.  By contrast, for $-4.25<\widehat{\pi}<-3.75$, the cost of
solving an instance is still negligible, but we cannot know in
advance whether it is solvable or not; here, $\pi=1/16$ implying that
the size-maximal biclique is still rather small, compared to the
cardinality of $V$ in the graph. Also, for $-3.75\geq\widehat{\pi}\geq-1.5$, the cost
of finding a biclique dramatically increases, but it is very likely that
a solution will be found. Notice an inflection point for $-1.5 < \widehat{\pi} < -1.2$ where it is negligible to find a biclique, after this point the computational cost increases significantly; this behaviour is more evident in Figs.~\ref{fig:percentileSize7}---\ref{fig:percentileSize8}. Lastly, notice that for $\widehat{\pi} > -1.2$ the computational cost decreases.

In conclusion, the phase transition for this problem exhibits an easy-hard-easy-hard-easy
pattern. Regardless of the parameter $z$ of the biclique being looked for, they all have similar
inflection points, but with different computational costs, as
expected. It also is worth noticing that the hardest instances lie
at the YES region: they are likely to be solvable. By way of
comparison, in the phase transition of many NP-complete problems they
lie at a no-decision region, where there is a 50\% probability of the
instance being solvable.

\subsection{Maximal Bicliques}
\label{sec:phaseoptimality}

Figs.~\ref{fig:percentileSize4} --- \Red{\ref{fig:percentileSize8}} result
from combining the output obtained for a number of problem instances,
some of which are solvable and other are not. To convey the behaviour
for either class separately, it is standard in the literature to find
optimal solutions to the problem at hand; then, set the parameter $z$ (to
be searched for) to be at a known `distance' $d\dfn\zmax-z$ from the
optimal solution for each problem, where $\zmax$ is the size of size-maximal 
biclique in a given instance. Hence, at $d=0, 1, 2, \ldots$
bicliques are guaranteed to be found, when $z\leq\zmax$;
indeed, at $d=0$ only size-maximal bicliques will be output.  However, at
$d=-1, -2, \ldots$ no bicliques will be found at all.

So, we have applied our algorithm to every problem instance, making it
find a size-maximal biclique. For this step, we slightly modified our
algorithm in two respects. First, we force it to report on the
computational cost involved in finding any biclique with $z\leq\zmax$,
because this implies solving this optimality problem
Second, we make the
algorithm to carry out a guarantee check, which rules out the
possibility for it to search for a biclique larger than the
maximum. Again, knowledge of the size-maximal biclique is readily
obtained from $\gram(\mathbf{Q})$.

We have found the size-maximal bicliques with maximal weight for the collection of graphs obtained by observing $50, 75, \ldots,
150$ actions from the real-world process. Interestingly, they all show the same phase transition pattern,
and, so, for the sake of brevity, we shall show only the mean cost to
find a solution.

\begin{figure}[ht!]
\centering
\includegraphics[scale=0.6]{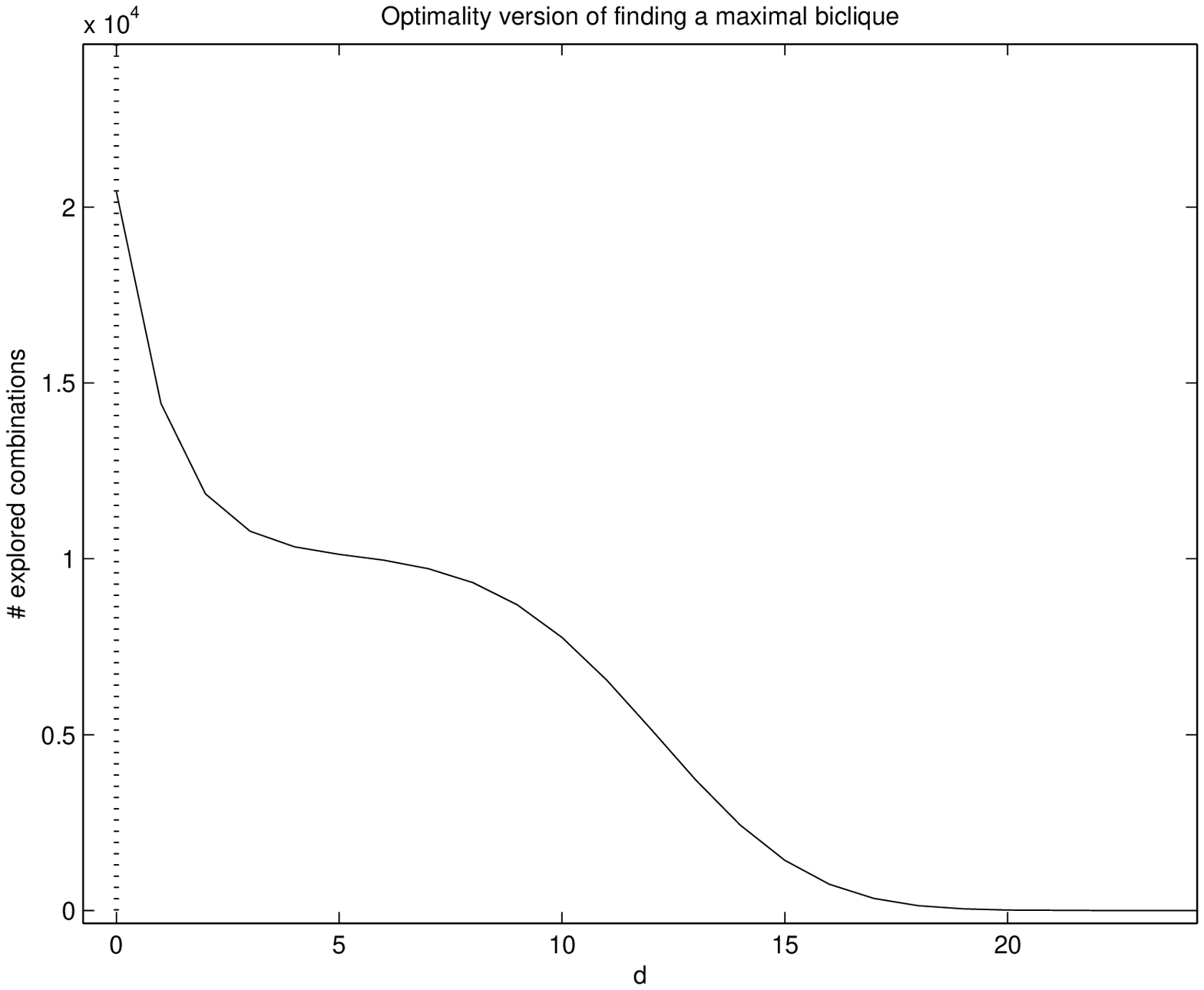}
\caption{Mean cost of finding a size-maximal biclique with maximal weight in a collection of graphs.}
\label{fig:meanOpt}
\end{figure}

Fig.~\ref{fig:meanOpt} shows the phase transition of finding the size-maximal biclique. We have set the biclique to be of maximal weight. In the curve, $d$ is the distance from the optimal solution. A solid line is used to denote the mean cost of finding a biclique with seize $d$; a dashed line to denote the boundary between solvable and unsolvable instances, after and before the dashed line, respectively.

Notice from Fig.~\ref{fig:meanOpt} that for the optimal solution, we have the highest computational cost of 20440 combinations explored. When we reach the optimal solution, for $0<d<3$, there is a significantly increase in the number of combinations explored. Then, for $3 \leq d\leq 10$ there is a soft increase in the number of combinations suggesting a critical area where the algorithm finds easy to discard combinations. By contrast, for $10<d< 15$ the computational cost increases significantly, this is because the algorithm finds difficult to discard combinations.

In conclusion, there is an exponential growth in the number of combinations as we reach the optimal solution.

\section{Conclusions and Indications for Further Work}
\label{sec:conclusions}

In this work we focused on the identification of the phase transition that separates easy and hard instances in the problem of finding a complete bipartite subgraph in a larger bipartite graph, namely a biclique. This problem naturally arises in several contexts (\eg~bioinformatics, social network analysis, document classification, etc.) however few efforts have been made towards understanding its phase transition. 

In order to conduct this work we followed a 4-step approach: First, the selection of the order parameter has been made by proposing a novel methodology which makes use of a classifier in order to identify parameter candidates. Second, in our experimentation we considered instances that arise in the context of a DNS server. Real-world instances can be significantly harder that similarly synthesised problems. Third, we have applied a backtracking based approach to solve problem instances and measure the computational expense involved. Lastly, we plot the computational cost, and the probability of an instance being solvable against the order parameter, denoted $\pi$ and given by the size of a size-maximal biclique divided over the cardinality of $V$.

Our results show a critical value at $\pi = 3/8$. There, any problem 
instance would be expensive in terms of the number of combinations 
to explore and with a 98\% probability of having a biclique with size $z$ and maximal weight.
By contrast, it takes no effort to realise that a graph does not contain a biclique for $\pi = 1/32$.

Our results help identifying key instances of the biclique problem that may be used to build a benchmark set, with which one can perform a statistically significant comparison amongst different methods that attempt to solve the same problem. Furthermore, the critical values of the phase transition study can also be used to determine when a complete method is likely to succeed (or not) in finding a solution to a problem instance in reasonable time.

In what follows we will describe some insights about future work: 1) It remains open how can we use the characteristics from the hardest instances to solve more efficiently the problem. As an example, consider that hardest instances have a common number of vertices $U$; then, it is possible to design an algorithm which makes use of a rule considering the number of vertices, so the algorithm may know in advance the computational cost involved in solving the instances. 2) It is possible to study this problem considering much larger graphs. However, our initial experimentation has shown that large graphs may take months to be completed with our algorithm. To address this, we are currently working on using a MapReduce approach following an approach similar as in~\cite{trejo}. 3) Our work considers only real-world instances, we can make use of random instances in an attempt to have a better insight of the phase transition of this problem. 4) It remains open whether our machine learning approach is suitable to identify order parameter candidates for other NP-complete problems, so further experimentation is necessary to determine the viability of this method.



 \bibliographystyle{elsarticle-num} 
 \bibliography{references}

\begin{thebibliography}{10}
\expandafter\ifx\csname url\endcsname\relax
  \def\url#1{\texttt{#1}}\fi
\expandafter\ifx\csname urlprefix\endcsname\relax\def\urlprefix{URL }\fi
\expandafter\ifx\csname href\endcsname\relax
  \def\href#1#2{#2} \def\path#1{#1}\fi

\bibitem{Cheeseman}
P.~Cheeseman, B.~Kanefsky, W.~M. Taylor, Where the {\em really} hard problems
  are, in: J.~Mylopoulos, R.~Reiter (Eds.), Proceedings of the 12th
  International Joint Conference on Artificial Intelligence, IJCAI, Morgan
  Kaufmann, 1991, pp. 331--337.

\bibitem{Gent95phasetransitions}
I.~P. Gent, T.~Walsh, Phase transitions from real computational problems, in:
  Proceedings of the 8th International Symposium on Artificial Intelligence,
  1995, pp. 356--364.

\bibitem{Hogg19961}
T.~Hogg, B.~A. Huberman, C.~P. Williams, Phase transitions and the search
  problem, Artificial Intelligence 81~(1--2) (1996) 1 -- 15, frontiers in
  Problem Solving: Phase Transitions and Complexity.

\bibitem{Mammen97anew}
D.~L. Mammen, T.~Hogg, A new look at the easy-hard-easy pattern of
  combinatorial search difficulty, Journal of Artificial Intelligence Research
  7 (1997) 47--66.

\bibitem{Frank}
J.~Frank, I.~P. Gent, T.~Walsh, Asymptotic and finite size parameters for phase
  transitions: Hamiltonian circuit as a case study, Information Processing
  Letters 65 (1998) 241--245.

\bibitem{Gent94thehardest}
I.~P. Gent, T.~Walsh, The hardest random {SAT} problems, in: B.~Nebel,
  L.~Dreschler-Fischer (Eds.), Proceedings of the 18th German Annual Conference
  on Artificial Intelligence, KI-94, Springer, 1994, pp. 355--366.

\bibitem{zhang}
Y.~Zhang, C.~Phillips, G.~Rogers, E.~Baker, E.~Chesler, M.~Langston, On finding
  bicliques in bipartite graphs: a novel algorithm and its application to the
  integration of diverse biological data types, BMC Bioinformatics 15~(1)
  (2014) 110.

\bibitem{Mushlin}
R.~Mushlin, A.~Kershenbaum, S.~Gallagher, T.~Rebbeck, A graph-theoretical
  approach for pattern discovery in epidemiological research, IBM Syst J 46
  (2007) 135--149.

\bibitem{Sanderson}
M.~Sanderson, A.~Driskell, R.~Ree, O.~Eulenstein, S.~Langley, Obtaining maximal
  concatenated phylogenetic data sets from large sequence databases, Mol Biol
  Evol 20~(7) (2003) 1036--1042.
\newblock \href {http://dx.doi.org/10.1093/molbev/msg115}
  {\path{doi:10.1093/molbev/msg115}}.

\bibitem{alonso-phd}
R.~Alonso, A social network based model to detect anomalies on {DNS} servers,
  Ph.D. thesis, Tecnol\'ogico de Monterrey (2015).

\bibitem{Yannakakis:1978:NEN:800133.804355}
M.~Yannakakis, Node-and edge-deletion np-complete problems, in: Proceedings of
  the Tenth Annual ACM Symposium on Theory of Computing, STOC '78, ACM, New
  York, NY, USA, 1978, pp. 253--264.

\bibitem{Dawande:2001:BMC:509217.509229}
M.~Dawande, P.~Keskinocak, J.~M. Swaminathan, S.~Tayur, On bipartite and
  multipartite clique problems, J. Algorithms 41~(2) (2001) 388--403.

\bibitem{Peeters2003651}
R.~Peeters, The maximum edge biclique problem is np-complete, Discrete Applied
  Mathematics 131~(3) (2003) 651 -- 654.

\bibitem{alonso2}
R.~Alonso, R.~Monroy, On the {NP}-completeness of computing the commonality
  amongst the objects upon which a collection of agents has performed an
  action, Computacion y Sistemas 17~(4) (2013) 489--500.

\bibitem{Gent96thetsp}
I.~P. Gent, T.~Walsh, The {TSP} phase transition, Artificial Intelligence
  88~(1--2) (1996) 349 -- 358.

\bibitem{Gent95scalingeffects}
I.~P. Gent, E.~MacIntyre, P.~Prosser, T.~Walsh, Scaling effects in the {CSP}
  phase transition, in: U.~Montanari, F.~Rossi (Eds.), Principles and Practice
  of Constraint Programming, CP '95, Springer, 1995, pp. 70--87.

\bibitem{Gent96phasetransitions}
I.~P. Gent, T.~Walsh, Phase transitions and annealed theories: Number
  partitioning as a case study, in: W.~Wahlster (Ed.), Proceedings of the
  Twelfth European Conference on Artificial Intelligence, ECAI'96, John Wiley
  \& Sons, 1996, pp. 170--174.

\bibitem{datamining}
I.~H. Witten, E.~Frank, Data Mining: Practical Machine Learning Tools and
  Techniques, 2nd Edition, Morgan Kaufmann Series in Data Management Systems,
  Morgan Kaufmann, 2005.

\bibitem{Gent94thesat}
I.~P. Gent, T.~Walsh, The {SAT} phase transition, in: A.~G. Cohn (Ed.),
  Proceedings of the Eleventh European Conference on Artificial Intelligence,
  ECAI'94, John Wiley \& Sons, 1994, pp. 105--109.

\bibitem{torres}
N.~Rangel-Valdez, J.~Torres-Jimenez, Phase transition in the bandwidth
  minimization problem, in: A.~Hern{\'a}ndez-Aguirre, R.~Monroy-Borja, C.~A.
  Reyes-Garc\'{\i}a (Eds.), Proceedings of the 8th Mexican International
  Conference on Artificial Intelligence, MICAI '09, Springer, 2009, pp.
  372--383.

\bibitem{trejo}
L.~Trejo, R.~Monroy, R.~Alonso, A.~Avila, M.~Maqueo, J.~Vazquez, E.~Sanchez,
  Using cloud computing {MapReduce} operations to detect {DDoS} attacks on
  {DNS} servers, in: A.~Proenca, A.~Pina, J.~Garc{\'\i}a-Tobio, L.~Ribeiro
  (Eds.), Proceedings of the 4th Iberian Grid Infrastructure Conference 2010,
  IBERGRID'10, Netbiblo, 2010, pp. 493--505.

\end{thebibliography}





\end{document}